\documentclass[letterpaper, 10 pt, conference]{ieeeconf}  % Comment this line out if you need a4paper

\IEEEoverridecommandlockouts                              % This command is only needed if 
\usepackage{multirow}
\usepackage{threeparttable} 
\usepackage{amsmath,amsfonts}
\usepackage{algorithmic}
\usepackage{algorithm}
\usepackage{array}
\usepackage[caption=false,font=normalsize,labelfont=sf,textfont=sf]{subfig}
\usepackage{textcomp}
\usepackage{stfloats}
\usepackage{url}
\usepackage{verbatim}
\usepackage{graphicx}
\usepackage{cite}
\usepackage{xcolor}
\usepackage{footmisc}
\makeatletter
\let\NAT@parse\undefined
\makeatother
\usepackage{hyperref}
\hypersetup{
    colorlinks=true,
    linkcolor=blue,
    citecolor=blue,
    urlcolor=blue,
    bookmarksnumbered=true
}
\usepackage{booktabs}
\usepackage{threeparttable}
\usepackage[numbers,sort&compress]{natbib}

\title{\LARGE \bf
Floorplan-SLAM: A Real-Time, High-Accuracy, and Long-Term Multi-Session Point-Plane SLAM for Efficient Floorplan Reconstruction
}

\author{Haolin Wang$^{1,2,\dagger}$, Zeren Lv$^{3,\dagger}$, Hao Wei$^{1,*}$, Haijiang Zhu$^{3}$ and Yihong Wu$^{1,2,*}$% <-this % stops a space
\thanks{*This work was supported by the National Natural Science Foundation of China under Grand No. 62402493 and the Youth Program of State Key Laboratory of Multimodal Artificial Intelligence Systems under Grand No. MAIS2024215. (Corresponding authors: Hao Wei and Yihong Wu.)}% <-this % stops a space
\thanks{$^{1,2}$Haolin Wang, Hao Wei and Yihong Wu are with the State Key Laboratory of Multimodal Artificial Intelligence Systems, Institute of Automation, Chinese Academy of Sciences, Beijing 100190, China. Haolin Wang and Yihong Wu are also with the School of Artificial Intelligence, University of Chinese Academy of Sciences, Beijing 100190, China (e-mail: \{wanghaolin2023; weihao2019; yihong.wu\}@ia.ac.cn).}
\thanks{$^{3}$Zeren Lv and Haijiang Zhu are with the College of Information Science and Technology, Beijing University of Chemical Technology, Beijing 100029, China (e-mail: 2022210463@buct.edu.cn, zhuhj@mail.buct.edu.cn).}%
\thanks{$^{\dagger}$Haolin Wang and Zeren Lv contributed equally to this work.}
}

\begin{document}

\maketitle
\thispagestyle{empty}
\pagestyle{empty}

\begin{abstract}
% We present Floorplan-SLAM, a real-time floorplan estimation algorithm that directly processes a sequence of stereo images without requiring active sensors. Unlike previous approaches, Floorplan-SLAM is designed for real-time sequential reconstruction, integrating pose estimation and floorplan reconstruction in a tightly coupled manner. Our system incrementally constructs plane landmarks using a novel stereo plane extraction method based on robust support points. We also introduce an innovative floorplan reconstruction strategy that dynamically selects an optimal subset of plane landmark segments based on the system state and scene structure constraints. Throughout the reconstruction process, both poses and plane landmarks are continuously optimized, enabling real-time floorplan reconstruction of complex indoor environments with high robustness, accuracy, and completeness.
% Additionally, our system supports long-term mapping, allowing reconstruction to extend existing floorplans in regions with overlapping areas without the need for redundant data collection.
% Experimental evaluations on the VECtor dataset and a self-collected dataset demonstrate that Floorplan-SLAM outperforms state-of-the-art methods that rely on a complete scene point cloud as input. It also surpasses existing stereo SLAM approaches in both reconstruction quality and pose estimation accuracy, achieving a real-time performance of 25-40 FPS.
Floorplan reconstruction provides structural priors essential for reliable indoor robot navigation and high-level scene understanding. However, existing approaches either require time-consuming offline processing with a complete map, or rely on expensive sensors and substantial computational resources. To address the problems, we propose Floorplan-SLAM, which incorporates floorplan reconstruction tightly into a multi-session SLAM system by seamlessly interacting with plane extraction, pose estimation, back-end optimization, and loop \& map merging, achieving real-time, high-accuracy, and long-term floorplan reconstruction using only a stereo camera. Specifically, we present a robust plane extraction algorithm that operates in a compact plane parameter space and leverages spatially complementary features to accurately detect planar structures, even in weakly textured scenes. Furthermore, we propose a floorplan reconstruction module tightly coupled with the SLAM system, which uses continuously optimized plane landmarks and poses to formulate and solve a novel optimization problem, thereby enabling real-time and high-accuracy floorplan reconstruction. Note that by leveraging the map merging capability of multi-session SLAM, our method supports long-term floorplan reconstruction across multiple sessions without redundant data collection. Experiments on the VECtor and the self-collected datasets indicate that Floorplan-SLAM significantly outperforms state-of-the-art methods in terms of plane extraction robustness, pose estimation accuracy, and floorplan reconstruction fidelity and speed, achieving real-time performance at 25--45 FPS without GPU acceleration, which reduces the floorplan reconstruction time for a 1000~$\mathrm{m}^{2}$ scene from 16 hours and 44 minutes to just 9.4 minutes.
% To the best of the authors’ knowledge, this is the first work that relies solely on a stereo camera and standard CPUs to achieve real-time floorplan reconstruction in large-scale indoor environments, making it well-suited for various robotics applications with strict cost and real-time requirements.
\end{abstract}

\section{INTRODUCTION}

% Floorplan reconstruction holds paramount importance for diverse downstream applications, including robot localization, indoor navigation, scene understanding. Indoor environments present unique challenges characterized by structural complexity, dense object clutter, and severe occlusion patterns that fundamentally limit visual perception. Existing methods rely on high-quality point clouds and are sensitive to point cloud noise, with reconstruction processes being complex and unable to operate in real-time. Despite three decades of active research, automated, real-time reconstruction of accurate floor plans through vision-only frameworks remains an open challenge, particularly in large-scale cluttered environments.
Floorplans are crucial for indoor robot navigation and high-level scene understanding, and are widely applied in scenarios ranging from service robots in assisted living environments to autonomous drones operating in complex indoor facilities. In these scenarios, generating a floorplan in real-time is particularly advantageous, as it not only provides geometric structure for accurate localization but also furnishes semantic cues for higher-level decision-making. Despite the growing need for real-time floorplan reconstruction, existing approaches often present significant limitations. Some methods operate in an offline manner, requiring a complete map of the environment prior to reconstruction \cite{han2021vectorized,nan2017polyfit,bauchet2020kinetic}. Others rely on expensive sensors such as LiDAR or RGB-D cameras \cite{solarte2022360,matez2022sigma}, or computationally intensive neural networks \cite{yue2023connecting,stekovic2021montefloor,chen2019floor}, which increase both deployment costs and system complexity. While monocular or stereo camera-based SLAM has attempted to use high-level features such as planes for real-time vectorized map construction \cite{yang2019monocular,wang2024rss,zhang2021stereo}, these methods lack high-level scene layout understanding, thereby failing to achieve complete and compact indoor vectorized reconstruction. Additionally, they cannot perform robust plane extraction in weakly textured scenes.

\begin{figure}[tp]
\centering
\includegraphics[width=\columnwidth]{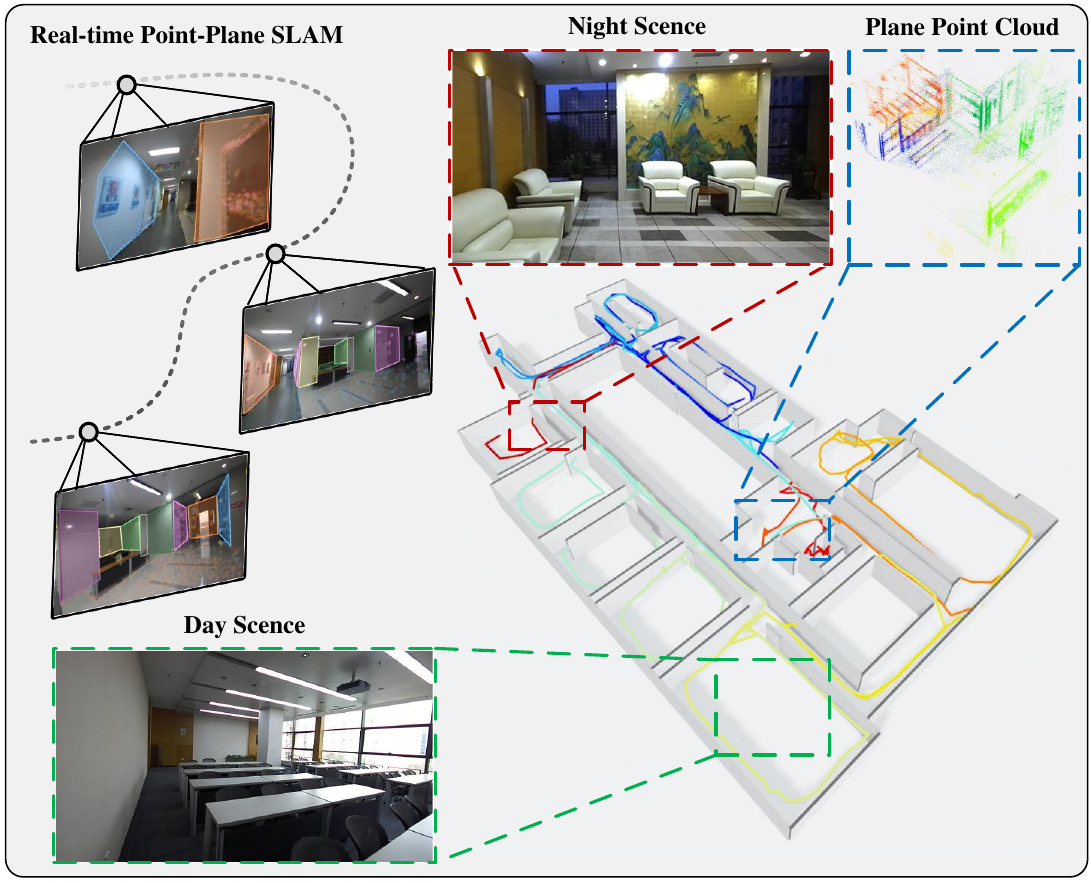}
\caption{Floorplan reconstruction results on the meeting rooms sequence from the self-collected dataset by Floorplan-SLAM. Floorplan-SLAM can robustly process stereo images containing low-texture and cluttered objects at 45 FPS on a CPU, and it merges multiple sub-sessions recorded at different times, ultimately achieving real-time, high-accuracy, and long-term floorplan reconstruction.}
\label{head_figure}
\end{figure}

To address the above problems, an effective real-time solution that works with more accessible and cost-effective sensors, such as stereo cameras, remains uncertain. In this work, we present Floorplan-SLAM, a novel framework that tightly integrates floorplan reconstruction into a multi-session SLAM system by seamlessly interacting with plane extraction, pose estimation, back-end optimization, and loop \& map merging. Unlike prior methods, our approach operates exclusively with stereo cameras under standard computing resources, allowing real-time performance in large-scale indoor settings. A robust plane extraction algorithm is first developed to handle weakly textured surfaces by leveraging two spatially complementary features in a compact plane parameter space. We further introduce a floorplan reconstruction module tightly integrated with the SLAM system that formulates and solves an innovative optimization problem using the continuously optimized plane landmarks and poses from the SLAM system, enabling real-time incremental floorplan reconstruction. Furthermore, by leveraging the map merging capability of multi-session SLAM, our method supports long-term floorplan reconstruction across multiple sessions without redundant data collection, making it more efficient for large-scale environments. To the best of the authors’ knowledge, this is the first work that relies solely on a stereo camera and standard CPUs to achieve real-time floorplan reconstruction in large-scale indoor environments, making it well-suited for various robotics applications with strict cost and real-time requirements.

The contributions of our work can be summarized as follows:

\begin{enumerate}
    \item We propose a robust plane extraction algorithm that operates in a compact plane parameter space and leverages two spatially complementary features to accurately and completely detect planar structures, even in weakly textured scenes.
    % \item We introduce a floorplan reconstruction module that is tightly coupled with the multi-session SLAM system, leveraging continuously optimized plane landmarks and poses from the SLAM system to formulate and solve a novel optimization problem, thereby enabling real-time incremental and long-term floorplan reconstruction.
    \item We introduce a floorplan reconstruction module that is tightly coupled with the multi-session SLAM system, leveraging continuously optimized plane landmarks and poses from the SLAM system to formulate and solve a novel binary linear programming problem with trajectory constraint thereby enabling real-time, high-accuracy, and long-term floorplan reconstruction.
    \item Experiments on the VECtor and our self-collected datasets demonstrate that Floorplan-SLAM significantly outperforms state-of-the-art methods in terms of plane extraction robustness, pose estimation accuracy, and floorplan reconstruction fidelity and speed, achieving real-time performance at 25--45 FPS without GPU acceleration.
\end{enumerate}

\section{Related Work}

Floorplan reconstruction aims to convert raw sensor data into vectorized geometric models. Some primitives-based vectorized reconstruction methods \cite{han2021vectorized,nan2017polyfit,bauchet2020kinetic} detect planes from the input point cloud and then perform a global optimization to select the optimal subset of candidate planes for vectorized reconstruction. However, these methods usually require a prior global map, such as a dense point cloud or a mesh map of the entire scene, which is difficult to obtain from actual point clouds or reconstructed dense meshes due to noise, outliers, and missing data.

With the development of neural networks, deep learning-based floorplan reconstruction methods have become of interest. These methods \cite{yue2023connecting,stekovic2021montefloor,chen2019floor} typically first project the input dense point cloud onto a top view to create a 2D density map, and then use neural networks to infer the floorplan of the scene from the projected density map. However, these methods are only applicable to specific scenarios, exhibit limited generalizability, and incur high time cost and computational resource consumption.

To achieve real-time floorplan reconstruction, some works \cite{yang2019monocular,solarte2022360,matez2022sigma} utilized visual SLAM to extract plane features from image sequences, which are then used to create landmarks for map construction. However, these methods either lack high-level scene layout understanding \cite{yang2019monocular}, thereby failing to achieve complete and compact indoor vectorized reconstruction, or integrate the acquired poses and scene structures in a loosely coupled manner \cite{solarte2022360,matez2022sigma}, leading to poor system robustness due to the inability to perform mutual optimization between pose and scene structure. In addition, these methods can only process one set of consecutive images at a time, making it impossible to merge multiple sequences and achieve long-term mapping.

Furthermore, these methods typically utilize RGB-D \cite{matez2022sigma} or neural networks \cite{yang2019monocular,solarte2022360} for plane extraction, which is challenging for stereo cameras. In \cite{zhang2021stereo}, a plane extraction method based on intersecting lines that satisfy specific geometric constraints was proposed. However, this method cannot accurately distinguish real planes from pseudo-planes and is highly susceptible to noise. To enhance the robustness of plane extraction, a region-growing-based plane extraction method was introduced in \cite{wang2024rss}. Nevertheless, it struggles to extract complete planes in weakly textured scenes and does not fully leverage visual information.

\begin{figure*}[tp]
\centering
\includegraphics[width=0.85\textwidth]{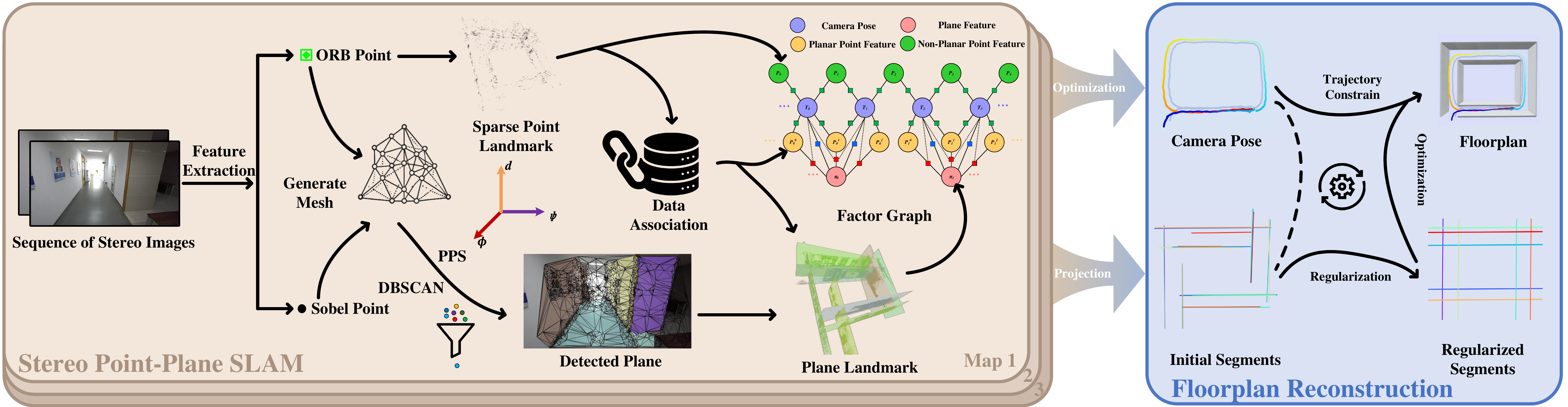}
\caption{System Overview. Floorplan-SLAM tightly integrates a stereo point-plane SLAM system with a floorplan reconstruction module. The former provides accurate plane landmarks and trajectory information, while the latter formulates and solves a novel optimization problem, ultimately achieving real-time, high-accuracy, and long-term floorplan reconstruction.}
\label{system_overview}
\end{figure*}

\section{Point-Plane-Based Stereo SLAM System}

% To transform and integrate the planar structure obtained from each frame into the global map, the camera pose of each frame needs to be estimated. Therefore, we propose a point-plane-based stereo SLAM system to estimate camera poses, while creating and managing plane landmarks, which are further processed for floorplan reconstruction in Section \ref{floorplanGeneration}. The proposed SLAM system utilizes point and plane features, as well as planar regularities, to estimate the camera pose. We refer the reader to \cite{wang2024rss} for more details on the pose estimation. Next, we focus on describing the operations related to plane landmark management and multi-session map merging.

In this section, we introduce the proposed point-plane-based stereo SLAM system. The system first uses two sets of spatially complementary point features to extract planes, and then employs both point and plane features to estimate poses. Plane landmarks are subsequently created and managed for floorplan reconstruction. In addition, a multi-session map merging mechanism is introduced to support long-term mapping. The overview of the system is illustrated on the left side of Fig. \ref{system_overview}. We refer the reader to \cite{wang2024rss} for more details on the pose estimation. Next, we focus on describing the operations related to plane extraction, plane landmark management, and multi-session map merging.

\subsection{Plane Extraction}

\subsubsection{Support Point Extraction}

We employ both the Sobel and ORB features for plane extraction, based on the insight that these two types of features focus on distinct yet complementary regions of a plane: Sobel predominantly captures boundary edges, while ORB detects interior corner points. By harnessing their complementary strengths, we achieve more robust and comprehensive plane extraction. Specifically, we first rectify the input stereo pair, ensuring that stereo correspondences are restricted to the same row in both images. Next, we perform ORB and Sobel feature matching across the stereo images to establish a set of correspondences, followed by consistency checks and outlier removal, thus obtaining a collection of 2D support points. For a 2D support point with pixel coordinates $ \left( u,v \right) $ and disparity value $d$, its 3D coordinates $ \mathbf{p}_s $ are computed through triangulation as follows:
\begin{equation}
\mathbf{p}_s=\left[ \begin{matrix}
	x&		y&		z\\
\end{matrix} \right] ^{\top}=\left[ \begin{matrix}
	\frac{\left( u-c_x \right) z}{f_x}&		\frac{\left( v-c_y \right) z}{f_y}&		\frac{f_xb}{d}\\
\end{matrix} \right] ^{\top},
\end{equation}
where $\left( f_x, f_y \right) $ is the focal length, $\left( c_x, c_y \right) $ is the principal point, and $b$ is the baseline, all known from calibration.

\subsubsection{Mesh Generation}

Because the support points acquired through stereo matching are relatively sparse, the estimated normal vectors of these points exhibit significant errors, rendering them unsuitable as reliable information for plane extraction. Therefore, we employ a set of triangles obtained via Delaunay triangulation \cite{chew1987constrained} on the 2D support points, with these support points serving as vertices, as the fundamental units for plane extraction. Then, we project these triangles into 3D space using the 3D coordinates of the support points to obtain a 3D mesh. We further prune the mesh by discarding triangles in the 3D mesh with long edges, high aspect ratios, and small acute angles \cite{rosinol2018densifying}.

\subsubsection{Plane Parameter Space Construction}

Each triangle in the 3D mesh is considered a planar patch, represented by $\pi =\left[ \mathbf{n}^{\top},d \right] ^{\top}$ in Cartesian space, where $\mathbf{n}=\left[ n_x,n_y,n_z \right] ^{\top}$ is the unit normal vector calculated by the cross product of two triangle edges, and $d$ is the distance from the origin to this triangle planar patch. We project these triangle planar patches from Cartesian space into Plane Parameter Space (PPS) \cite{sun2017rgb}, which compactly represents planes in Cartesian space. Specifically, given a triangle planar patch in Cartesian space, its representation in the PPS is a point $ \mathbf{p}_{\pi}$:
\begin{equation} \label{pps_representation}
\mathbf{p}_{\boldsymbol{\pi }}=\left[ \begin{array}{l}
	\phi\\
	\psi\\
	d\\
\end{array} \right] =\left[ \begin{array}{c}
	\mathrm{atan2}\left( n_y,n_x \right)\\
	\mathrm{arccos} \left( n_z \right)\\
	d\\
\end{array} \right] ,
\end{equation}
where $ \phi$ and $\psi$ are the azimuth and elevation angles of the normal vector respectively.

\subsubsection{Plane Parameters Estimation}

Ideally, the coplanar triangle planar patches in Cartesian space share the same coordinates in the PPS. However, due to noise, the coordinates of these coplanar triangle planar patches in the PPS are not exactly the same but rather close. Therefore, the problem of detecting coplanar triangle planar patches in Cartesian space is transformed into a point clustering problem in the PPS. We employ DBSCAN \cite{ester1996density} to cluster the points in the PPS, which is robust to noise.

After obtaining the coplanar triangle planar patches, the vertices of these triangles are coplanar points. We apply RANSAC \cite{fischler1981random} to these coplanar points to obtain accurate plane parameters, and only planes with an inlier ratio exceeding $ \theta _i $ are added to the plane feature set. In our experiments, $ \theta _i $ is set to 0.75.

\subsection{Plane Landmark Management}

We create new plane landmarks using planes extracted from keyframes that do not match any existing plane landmarks in the map, where planes are considered matched if the angle between their normal vectors is less than $\theta_{m}$ and the minimum distance from a planar point to the plane is less than $d_{m}$. These newly created plane landmarks are initially marked as invalid due to their potential unreliability, and become valid once they have been observed by more than thirty frames and three keyframes. Only valid plane landmarks are utilized for pose estimation and floorplan reconstruction. However, if a valid plane landmark is observed by at most one keyframe due to excessive optimization errors, it is reverted to invalid. For any two plane landmarks $ \mathbf{\Pi }_i$ and $ \mathbf{\Pi }_j$, if the angle between their normal vectors is less than $\theta _m$ and the minimum point-to-plane distances from the support points of $\mathbf{\Pi}_i$ to $\mathbf{\Pi}_j$ and from $\mathbf{\Pi}_j$ to $\mathbf{\Pi}_i$ are both less than $d_m$, we remove the plane landmark with fewer keyframe observations and transfer its observation information and support points to another plane landmark, followed by RANSAC to update the plane points. The above operations continue throughout the entire process of the system. In our experiments, $\theta _m$ is set to 5$^\circ$, and $d_m$ is set to 2 cm.

\subsection{Multi-Session Map Merging}

We use DBoW2 \cite{galvez2012bags} for place recognition and compute the aligning transformation $\mathbf{T}_{cm}$ between the two maps. The points and plane landmarks from the current map are transformed into the matching map using $\mathbf{T}_{cm}$. Duplicated landmarks from the current map are removed, and their observations are transferred to the corresponding landmarks in the matching map. We then perform local bundle adjustment (BA), pose graph optimization, and global BA in sequence, similar to \cite{campos2021orb}, to optimize the merged map and enhance overall consistency. By merging sequences captured at different times, we do not need to capture all map areas in a single pass, which is crucial for large-scale scene long-term mapping, as it significantly reduces the computational and storage burden on the acquisition devices. Notably, during each BA process, the parameters of plane landmark $\mathbf{\Pi }_k$ are optimized, and its plane points are updated by aggregating the support points of its observations $\boldsymbol{\pi }_{i,k}$ in the keyframes into $\mathbf{\Pi }_k$ followed by RANSAC, thereby providing a more accurate scene layout for the subsequent floorplan reconstruction.

\section{Floorplan Reconstruction} \label{floorplanGeneration}

The floorplan reconstruction module runs in parallel with the tracking, local mapping, and loop \& map merging threads. It processes the plane landmarks and trajectory information generated by the SLAM system and reconstructs the floorplan by formulating and solving a novel optimization problem. An overview of this workflow is shown on the right side of Fig. \ref{system_overview}.

\subsection{Candidate Wall Segment Generation}

We first select valid plane landmarks that are approximately perpendicular to the ground. We then project their plane points onto the ground to obtain projection lines and support points, thereby generating 2D wall segments. We further merge segments whose angle is less than $\theta_{r}$ and share more than $n_r$ support points sufficiently close to both segments, in order to improve the regularity of the wall segments. Next, we extend and pairwise intersect these segments to generate candidate wall segments. In our experiments, $\theta_{r}$ is set to 10$^\circ$, and $n_r=\min \left( \left| \mathrm{SP}\left( s_i \right) \right|,\left| \mathrm{SP}\left( s_j \right) \right| \right) /10$, where $\left| \mathrm{SP}\left( s_i \right) \right|$ represents the number of support points of $s_i$.

\subsection{Wall Segment Selection}

% The pairwise intersection process in the previous step introduces redundant candidate wall segments. Therefore, given a set of candidate wall segments $C=\left\{ c_1,c_2,\dots ,c_N \right\} $, we aim to select an optimal subset of this set through global optimization, thereby achieving accurate floorplan reconstruction.

% \paragraph{Problem Formulation}

% We formulate the optimization problem as a binary linear programming problem under hard constraints. Specifically, we introduce binary variables $x_i,i\in \left\{ 1,2,\dots ,N \right\} $, where $x_i = 1$ indicates that $c_i$ is selected, and $x_i = 0$ indicates that $c_i$ is not selected. The objective function of this optimization problem consists of three energy terms, including the point fitting term $E_f$, point coverage term $E_c$, and model complexity term $E_m$. Furthermore, multiple hard constraints are formulated to ensure that the reconstructed floorplan is initially closed, and on this basis, further includes openings. The optimization problem is solved through an energy minimization approach, and the floorplan is reconstructed by assembling the candidate wall segments with $x_i = 1$.

Given a set of candidate wall segments $C=\left\{ c_1,c_2,\dots ,c_N \right\}$ introduced by the pairwise intersection process, we aim to select an optimal subset of this set through global optimization, thereby achieving accurate floorplan reconstruction. We formulate the optimization problem as a binary linear programming problem under hard constraints and solve it via an energy minimization approach. The objective function of this optimization problem consists of three energy terms, the point fitting term $E_f$, the point coverage term $E_c$, and the model complexity term $E_m$, subject to multiple hard constraints.

\subsubsection{Point Fitting Term}

This term aims to evaluate the fitting degree between candidate wall segments and their support point sets:
\begin{equation}
E_f=1-\frac{1}{\left| P \right|}\sum_{i=1}^N{f\left( c_i \right) x_i},
\end{equation}
\begin{equation}
f\left( c \right) =\sum_{p\in \mathrm{SP}\left( c \right) \left| \mathrm{dist}\left( p,c \right) <\varepsilon _f \right.}{\left( 1-\frac{\mathrm{dist}\left( p,c \right)}{\varepsilon _f} \right)},
\end{equation}
where $ \left| P \right| $ is the total number of support points of candidate wall segments, $\mathrm{SP}\left( c \right) $ is the support point set of segment $c$, and $\mathrm{dist}\left( p,c \right)$ is the distance from point $p$ to segment $c$. Only points with a distance less than $\varepsilon _f$ to the corresponding segment are used in the calculation of $f\left( c \right)$, and $\varepsilon _f$ is set to the average distance between candidate wall segments and their support point sets in our experiment.

% Intuitively, the greater the number of support points of $c_i$ and the closer they are to $c_i$, the larger the value of $ f\left( c \right) $ will be. In order to minimize the value of $E_f$ as much as possible, the probability of $c_i$ being selected increases accordingly.

\subsubsection{Point Coverage Term}

% This term aims to appropriately evaluate the coverage degree between candidate wall segments and their support point sets in cases where the point cloud is incomplete due to missing data. We project the support points of candidate wall segment $c_i$ onto $c_i$ and calculate the distances between adjacent projected points. If the distance is less than $\varepsilon_c$, the segment between the two projected points is considered to be covered by support points:

This term aims to appropriately evaluate the coverage degree between candidate wall segments and their support point sets in cases where the point cloud is incomplete due to missing data. We consider a segment to be covered by two adjacent support points if the distance between these points is less than $\varepsilon_c$:
\begin{equation}
E_c=\frac{1}{N}\sum_{i=1}^N{\left( 1-\frac{\mathrm{len}_{\mathrm{cov}}\left( c_i \right)}{\mathrm{len}\left( c_i \right)} \right) x_i},
\end{equation}
where $\mathrm{len}_{\mathrm{cov}}\left( c_i \right)$ is the total length of the portion of $c_i$ covered by its support points, and $\mathrm{len}\left( c_i \right) $ is the length of $c_i$. In our experiments, $\varepsilon _c=10\cdot \mathrm{density}\left( P \right)$, where $\mathrm{density}\left( P \right)$ represents the average of the mean distances between all support points and their respective 10-nearest neighbors.
% Intuitively, this term tends to select candidate wall segments whose support points densely and uniformly cover them.

\subsubsection{Model Complexity Term}

% Due to noise, outliers, and missing data of point clouds, some sharp structures are erroneously introduced into the floorplan, and this term aims to evaluate the complexity of the reconstructed floorplan. Specifically, given the set of intersection points $ V=\left\{ v_1,v_2,\dots ,v_M \right\} $ generated by pairwise intersection of the wall segments, if an intersection point $v_i$ is connected by non-collinear candidate wall segments $c_i$ and $c_j$, and both $c_i$ and $c_j$ are selected, we consider $v_i$ to introduce a sharp structure:
This term aims to evaluate the complexity of the reconstructed floorplan and maintain it at an appropriate level of complexity. Specifically, if an intersection point $v_i$ generated by pairwise intersection of the wall segments is connected by non-collinear candidate wall segments $c_i$ and $c_j$, and both $c_i$ and $c_j$ are selected, we consider $v_i$ to introduce a sharp structure:
\begin{equation}
E_m=\frac{1}{M}\sum_{i=1}^M{\mathbb{I}_{\mathrm{sharp}} \left( v_i \right)},
\end{equation}
where $\mathbb{I}_{\mathrm{sharp}} \left( v_i \right)$ is an indicator function that equals 1 if $v_i$ introduces a sharp structure, and 0 otherwise. 
% Intuitively, this term encourages simple structures, i.e., intersection points are connected by collinear segments whenever possible, thereby maintaining an appropriate level of complexity for the reconstructed floorplan.

\begin{figure}[tp]
\centering
\includegraphics[width=0.85\columnwidth]{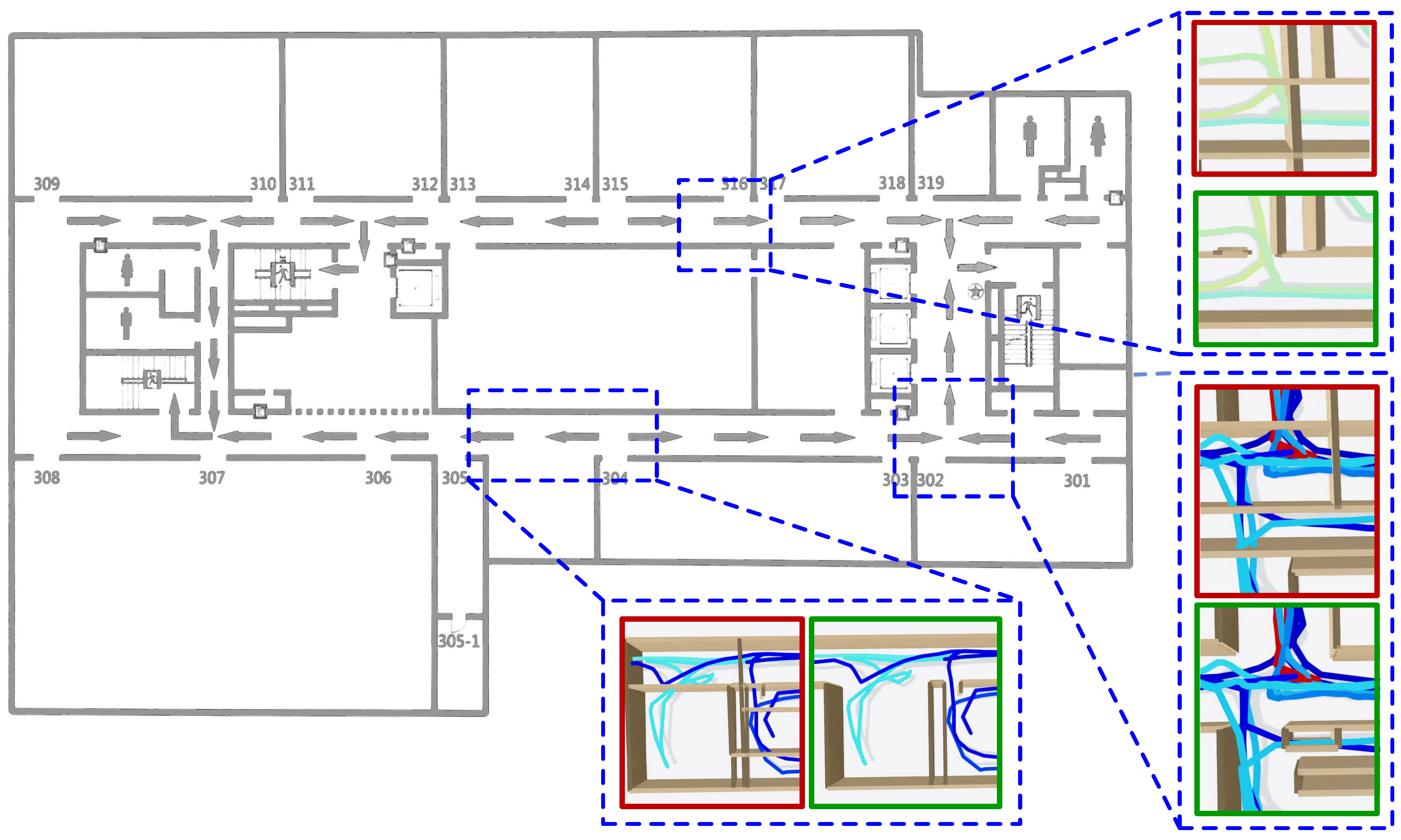}
\caption{Comparison of floorplan reconstruction results with and without trajectory constraints. The top-left shows the CAD floor plans provided by the meeting room scenario in the self-collected dataset. The red box indicates the reconstruction results without trajectory constraints, while the green box represents the reconstruction results with trajectory constraints. The trajectory constraints ensure that the reconstructed floorplan contains navigable openings, thereby significantly improving reconstruction accuracy.}
\label{traj_constraints}
\end{figure}

\subsubsection{Constraints}

% These constraints aim to ensure the reconstructed floorplan is closed except at openings. Trajectory information is a unique component of the SLAM system, as it inherently contains prior knowledge about navigable areas. We first utilize trajectory information to detect openings, as shown in Fig. \ref{traj_constraints}. If a candidate wall segment is crossed by the trajectory, it is considered an opening and is excluded from selection:
Trajectory information is a unique component of the SLAM system, as it inherently contains prior knowledge about navigable areas, as shown in Fig. \ref{traj_constraints}. We exclude candidate wall segments that are crossed by the trajectory from the selection:
\begin{equation}\label{constraint_trajectory}
\sum_{i=1}^{N} \mathbb{I}_{\mathrm{cross}} \left ( c_{i} \right ) \cdot x_{i} = 0,
\end{equation}
where $\mathbb{I} _{\mathrm{cross}} \left( c_i \right) $ is an indicator function that equals 1 if $ c_i $ is crossed by the trajectory, and 0 otherwise. To ensure that the floorplan is closed except at the passable areas, the intersection point $v_i$, which is not connected by segments crossed by the trajectory, must be connected by at least two and at most four candidate wall segments. Otherwise, it must be connected by at least one and at most $ 4-n_c $ candidate wall segments:
\begin{equation}\label{constraint_openings}
\left\{
\begin{aligned}
    \sum_{i=1}^N{x_i} &\in \left\{ 0, 2, 3, 4 \right\}, & \text{if} \,\, \forall c_i \in \mathcal{C} \left( v_j \right), \, \mathbb{I} \left( c_i \right) = 0, \\
    0 \le \sum_{i=1}^N{x_i} &\le 4 - n_c, & \text{if} \,\, \exists c_i \in \mathcal{C} \left( v_j \right), \, \mathbb{I} \left( c_i \right) = 1,
\end{aligned}
\right.
\end{equation}
where $\mathcal{C} \left( v_j \right)$ and $n_c$ represent the set of candidate wall segments connected to $v_j$ and the number of segments in the set that are traversed by the trajectory, respectively.

\subsubsection{Optimization}

The objective function $E$ of this optimization problem is the weighted sum of the above energy terms:
\begin{equation}\label{objective_function}
E=\lambda _fE_f+\lambda _cE_c+\lambda _mE_m.
\end{equation}
We utilize the SCIP solver \cite{gamrath2020scip} to minimize (\ref{objective_function}) subject to the hard constraints (\ref{constraint_trajectory}) and (\ref{constraint_openings}) to obtain the optimal subset of candidate wall segments. Then, the floorplan is reconstructed by assembling the segments in the optimal subset.

\section{Experiment}

\begin{figure}[tp]
\centering
\includegraphics[width=0.5\columnwidth]{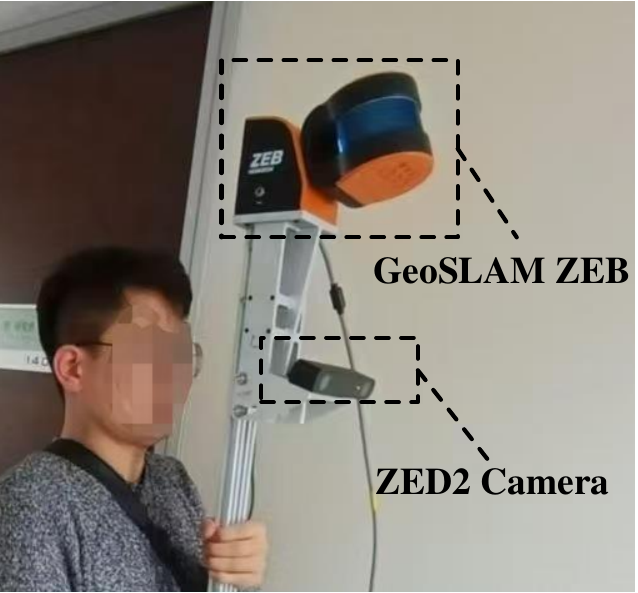}
\caption{Our self-collected dataset setup.}
\label{setup}
\end{figure}

We evaluated our method on the public VECtor \cite{gao2022vector} dataset and a self-collected dataset. The self-collected dataset was captured using a ZED2 camera with a GeoSLAM ZEB as ground truth, as shown in Fig. \ref{setup}. All experiments were carried out on a desktop with an AMD Ryzen 7 4800H CPU (running at 2.9 GHz) and 32 GB of RAM.

%\subsection{Experiment Setup} \label{section Experiment Setup}
%We evaluate the proposed Floorplan-SLAM on two datasets: VECtor and a self-collected dataset. The VECtor dataset captures various large-scale indoor environments with long trajectories and weak texture, consisting of six sequences across three scene types (corridors, schools, and residential units). The average trajectory length for each sequence exceeds 100 meters. Notably, the VECtor dataset does not provide point cloud ground truth, so we utilize a state-of-the-art LiDAR-SLAM method to generate ground truth point clouds.
%The self-collected dataset is recorded using our setup (Fig. \ref{setup}). We collect stereo images with a ZED2 camera, while GeoSLAM results serve as the ground truth for both camera poses and point clouds. This dataset consists of seven sequences, which feature rich textures but also contain significant occlusions from objects such as furniture, tables, and chairs. Four sequences (Corridor, Room, Stairs, and Parking Lot) are used for localization accuracy evaluation, whereas the remaining three sequences (Floor3, Floor14, and Floor17), which only have point cloud ground truth, are used for floorplan reconstruction evaluation.

\subsection{Plane Extraction Performance Evaluation} \label{plane_extraction_performance_evaluation}

Since we utilize the extracted planes for floorplan reconstruction, the accuracy and stability of these planes are crucial to the overall quality of the reconstruction outcomes. We evaluate our plane extraction algorithm and compare it with the plane extraction modules of two point-plane-based stereo SLAM systems, Stereo-Plane-SLAM \cite{zhang2021stereo} and RSS \cite{wang2024rss}. We use plane reprojection error and plane observation count as evaluation metrics to measure the accuracy and stability of the extracted planes, respectively. Specifically, the plane reprojection error represents the matching error between plane observations and landmarks, while the plane observation count indicates the average number of times each valid plane landmark is observed by keyframes. To eliminate the impact of trajectory errors on plane extraction performance, we utilize ground-truth trajectories for evaluation.

\begin{figure}[tp]
\centering
\includegraphics[width=\columnwidth]{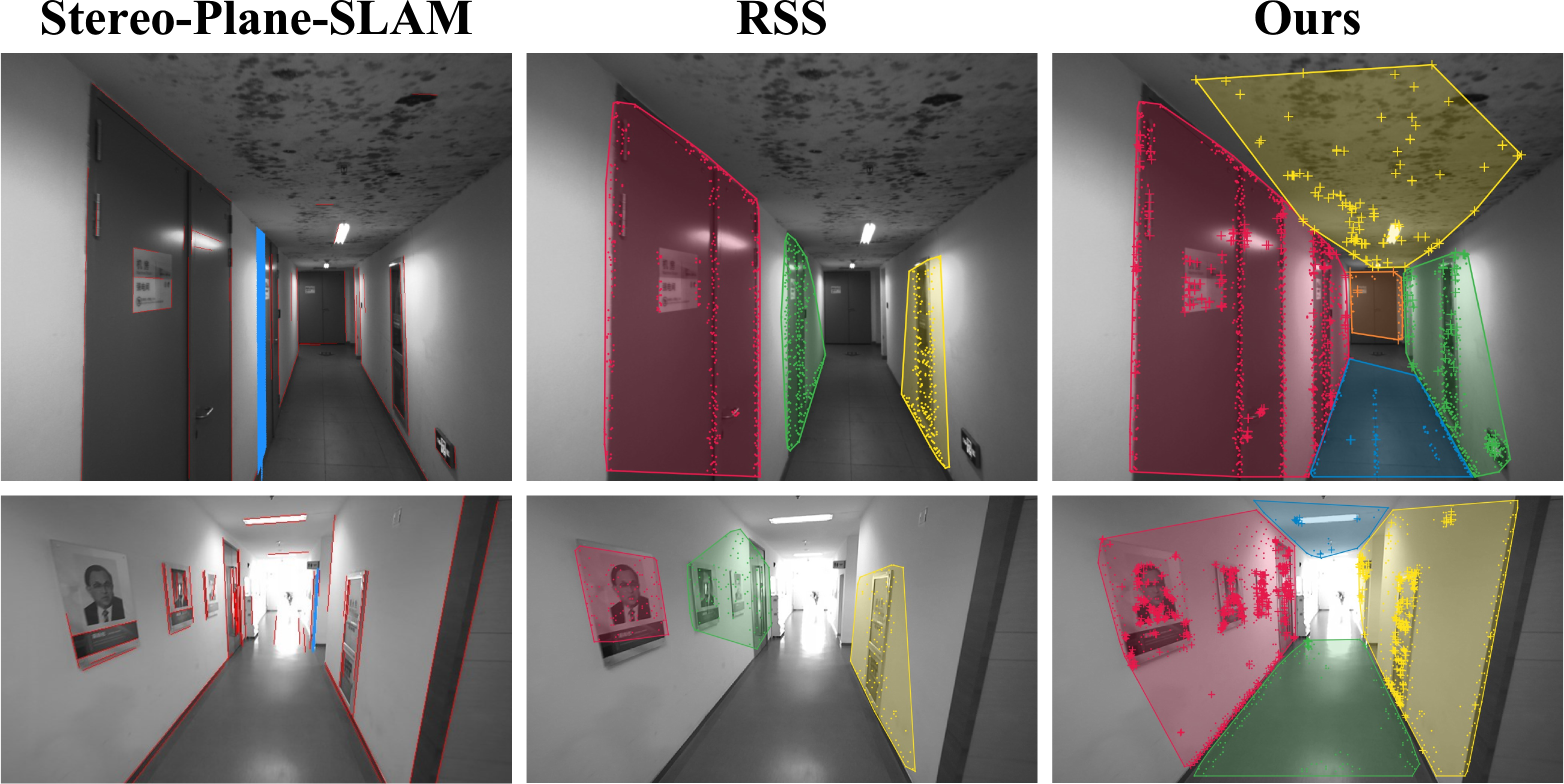}
\caption{Comparison of plane extraction results on the VECtor (upper) and self-collected (lower) datasets. ORB features are represented by crosses, and Sobel features are represented by dots in our method. The plane boundaries in both RSS and our approach are determined by the convex hull of the corresponding plane points.}
\label{plane_extraction_qualitative}
\end{figure}

Table \ref{plane_extraction_quantitative} compares the plane extraction performance of different methods on the VECtor and self-collected datasets. Our method significantly outperforms SP-SLAM and RSS in both accuracy and stability. Because RSS uses a region-growing strategy for plane extraction, insufficient support points in weakly textured regions can lead to multiple planes being extracted from a single surface, making them more prone to noise and thus less stable. In contrast, our method employs DBSCAN clustering in the PPS space to robustly detect complete and accurate discontinuous planes even in weakly textured scenes lacking support points. Additionally, by using both ORB and Sobel features, which focus on corner and edge regions respectively, our method can extract more planes than RSS, which relies solely on Sobel features, particularly in VECtor scenarios with numerous patches on ceilings and floors, as illustrated in Fig. \ref{plane_extraction_qualitative}. SP-SLAM relies on intersecting lines for plane extraction. Although many line segments can be identified, strict filtering to avoid spurious planes leaves only a few that contribute to plane extraction. Moreover, these segments are highly sensitive to noise, making the extracted planes unstable across multiple frames and reducing their overall accuracy and stability.

\begin{table}[]
\centering
\begin{threeparttable}
\caption{Plane Extraction Performance Comparison on the VECtor and Self-Collected Datasets}
\label{plane_extraction_quantitative}
% \resizebox{\columnwidth}{!}{%
\begin{tabular}{@{}l|ccc|ccc@{}}
\toprule
\multirow{2}{*}{Sequence} & \multicolumn{3}{c|}{Reprojection Error $\downarrow$} & \multicolumn{3}{c}{Observation Count $\uparrow$} \\
                          & SP\textsuperscript{1}       & RSS      & Ours              & SP\textsuperscript{1}      & RSS     & Ours              \\ \midrule
corridors                 & 0.033    & 0.028    & \textbf{0.009}    & 11.909  & 32.059  & \textbf{103.727}  \\
school                    & 0.039    & 0.029    & \textbf{0.011}    & 6.250   & 28.265  & \textbf{79.375}   \\ \midrule
meeting rooms     & 0.048    & 0.017    & \textbf{0.007}    & 15.333  & 34.867  & \textbf{226.286}  \\
offices                   & 0.062    & 0.022    & \textbf{0.008}    & 19.750  & 22.528  & \textbf{135.167}  \\
café              & 0.057    & 0.024    & \textbf{0.008}    & 14.333  & 20.286  & \textbf{147.250}  \\ \bottomrule
\end{tabular}%
\begin{tablenotes}
\footnotesize
\item[1] SP is the abbreviation for Stereo-Plane-SLAM \cite{zhang2021stereo}.
\end{tablenotes}
\end{threeparttable}
% }
\end{table}

\subsection{Localization Accuracy Evaluation}

% Accurate poses ensure that the local planes extracted from different frames can be seamlessly integrated into the global map, thereby maintaining the global consistency of the incrementally reconstructed floorplan. We evaluate the localization performance of our method in large-scale indoor scenes and compare it with the most relevant state-of-the-art visual SLAM systems, including ORB-SLAM3 (in stereo mode) \cite{campos2021orb}, Stereo-Plane-SLAM \cite{zhang2021stereo}, and RSS \cite{wang2024rss}. To evaluate the effectiveness of the plane constraint introduced by the plane features, we also conducted an ablation study with a point-only variant of our system. We conduct experiments on the public dataset VECtor \cite{gao2022vector} and a self-collected dataset, both of which contain several sequences with trajectory lengths exceeding 100 meters. The self-collected dataset, serving as the target scenario for the subsequent floorplan reconstruction experiment, was captured using a ZED2 camera, and the GeoSLAM results are employed as the ground-truth trajectory. The acquisition setup is shown in Fig. \ref{setup}. We align the estimated trajectory and the ground truth using the Umeyama algorithm without scaling \cite{umeyama1991least} and use the root mean square error (RMSE) of the absolute trajectory error (ATE) \cite{sturm2012benchmark} as the evaluation metric to compare global drift. Considering the randomness introduced by the multi-threaded system, we run each sequence ten times and report the median results.

Accurate poses allow planes extracted from different frames to be seamlessly integrated into the global map, maintaining the global consistency of the incrementally reconstructed floorplan. We evaluate the localization performance of our method in large-scale indoor scenes and compare it with the most relevant state-of-the-art visual SLAM systems, including ORB-SLAM3 (stereo mode) \cite{campos2021orb}, Stereo-Plane-SLAM \cite{zhang2021stereo}, and RSS \cite{wang2024rss}. To assess the impact of plane features, we also conduct an ablation study using a point-only variant of our system. Experiments are performed on the public VECtor dataset \cite{gao2022vector} and the self-collected dataset, both containing sequences several hundred meters in length. We disable the loop closure module and align the estimated trajectory with ground truth using the Umeyama algorithm without scaling. The RMSE of the absolute trajectory error (ATE) is used to measure global drift. Each sequence is run ten times to mitigate multi-thread randomness, and the median results are reported.

\begin{table}[]
\centering
% % \resizebox{0.8\columnwidth}{!}{%
\begin{threeparttable}
\caption{Localization Performance Comparison on the VECtor and Self-Collected Dataset (ATE RMSE {[}m{]})}
\label{localization_performance_evaluation}
\begin{tabular}{@{}l|ccccc@{}}
\toprule
Sequence         & ORB\textsuperscript{1}  & SP\textsuperscript{1}            & RSS  & PO\textsuperscript{1}    & Ours          \\ \midrule
corridors\_dolly & 0.96 & 0.95          & 0.94 & 0.96 & \textbf{0.93} \\
units\_dolly     & 2.48 & 2.49          & 1.81 & 2.42 & \textbf{1.64} \\
units\_scooter   & 1.89 & 1.82          & 1.74 & 1.84 & \textbf{1.41} \\
school\_dolly    & 1.40 & 1.39          & 1.40 & 1.39 & \textbf{1.37} \\
school\_scooter  & 1.39 & \textbf{1.30} & 1.33 & 1.38 & 1.31          \\ \midrule
café         & 1.53 & 1.57          & 1.38 & 1.54 & \textbf{1.29} \\
offices             & 1.40 & 1.25          & 1.18 & 1.38 & \textbf{1.07} \\
meeting rooms            & 2.35 & 1.79          & 1.66 & 2.40 & \textbf{1.60} \\
basement         & 2.49 & 2.41          & 1.61 & 2.52 & \textbf{1.56} \\ \bottomrule
\end{tabular}
% }
\begin{tablenotes}
\footnotesize
\item[1] ORB, SP, and PO are abbreviations for ORB-SLAM3, Stereo-Plane-SLAM, and the point-only variant of our system, respectively.
\end{tablenotes}
\end{threeparttable}
% }
\end{table}

As shown in Table \ref{localization_performance_evaluation}, our method outperforms ORB-SLAM3 and the Point-Only variant across all sequences, especially on the units sequences of the VECtor dataset and the basement sequence of the self-collected dataset, which feature challenging scenes characterized by low texture, dynamically changing illumination, and motion blur. In these scenarios, our approach achieves a notable accuracy improvement of 0.5$\,$--$\,$0.9$\,$m. This can be attributed to two primary reasons: (1) Compared with point correspondences, plane correspondences are more accurate and stable, particularly in the aforementioned challenging scenarios; (2) Planar structures, especially life-long structures such as walls with extensive spatial coverage, can be tracked across considerable distances, thereby providing comprehensive and reliable constraints for pose estimation. Furthermore, in comparison with the two point-plane-based stereo SLAM systems, our method yields the smallest average RMSE and still demonstrates superior performance on these challenging sequences. This can be attributed to the more comprehensive and accurate plane constraints introduced by our advanced plane extraction method, as demonstrated in Section \ref{plane_extraction_performance_evaluation}.

% \subsection{Floorplan Reconstruction Performance Evaluation}
\begin{figure}[tp]
\centering
\includegraphics[width=\columnwidth]{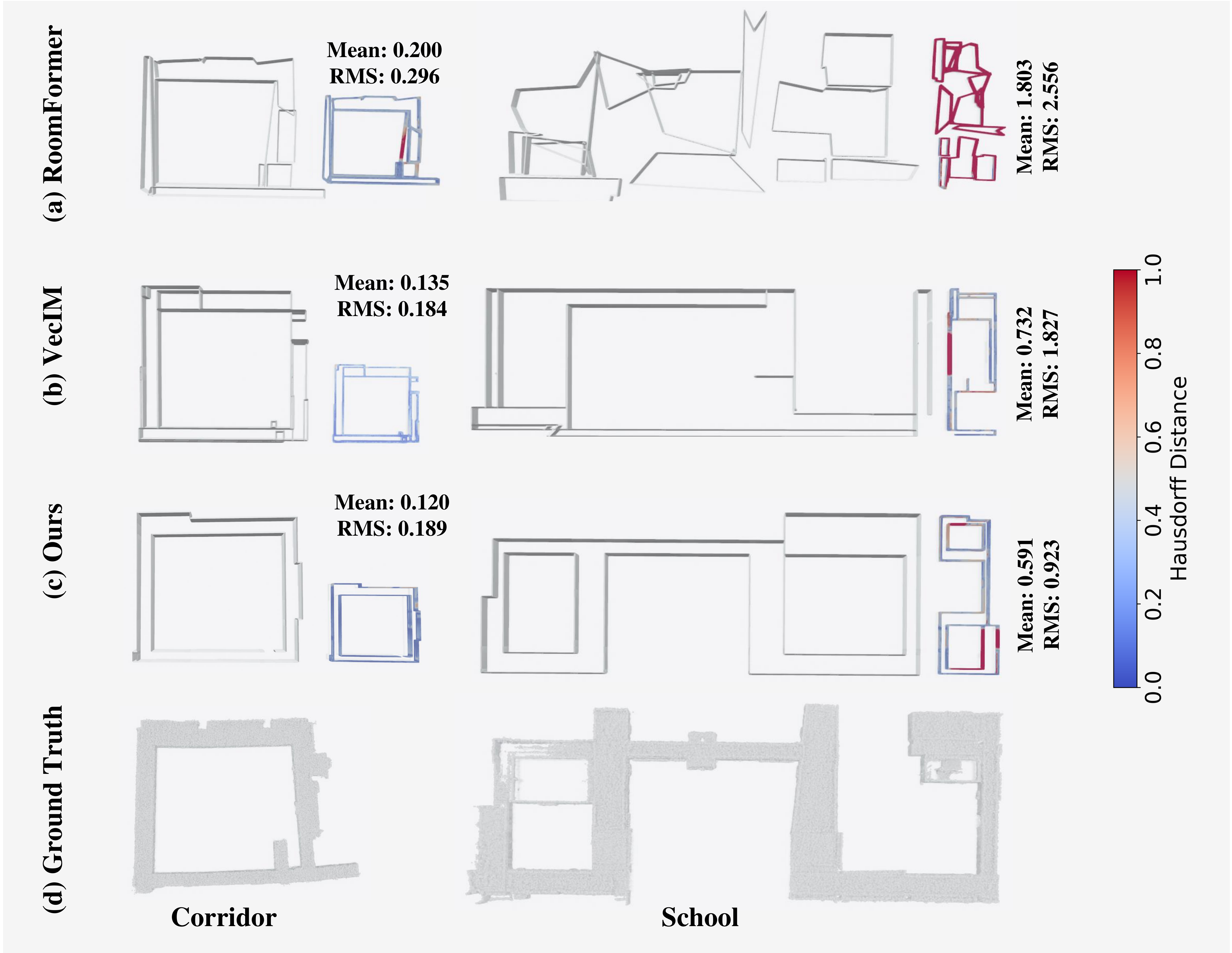}
\caption{Floorplan reconstruction results on the VECtor dataset of RoomFormer, VecIM, and Ours. We compute and visualize the mean and RMS Hausdorff error (m) between the reconstructed models and the LiDAR point cloud ground truth. The error distribution is color-coded, where blue indicates smaller errors and red represents larger errors. %We performed scale alignment on the results of the Corridor sequence.
}
\label{Vector_exp}
\end{figure}

\subsection{Floorplan Reconstruction Performance Evaluation}

% In this section, we evaluate the 3D floorplan models generated by our proposed method on two datasets, VECtor and Self-collect, which exhibit different characteristics and levels of complexity. We compare our results with two state-of-the-art methods: VecIM, a general vectorized reconstruction approach, and RoomFormer, a learning-based method.
% VecIM takes a point cloud aligned with gravity as input, whereas RoomFormer processes a density map. For both the VECtor and Self-collect datasets, we generate multi-view stereo (MVS) point clouds using COLMAP to serve as input for VecIM and RoomFormer. In contrast, our method requires only the original stereo images as input.
% Since these datasets do not provide ground-truth vectorized models, we use dense, high-precision LiDAR point clouds as the ground truth for quantitative evaluation. To assess model accuracy, we uniformly sample 10,000 points from the reconstructed models and compute the Hausdorff distance between the sampled points and the ground-truth point clouds.

We evaluate the floorplan reconstruction performance of our method on the VECtor \cite{han2021vectorized} and the self-collected dataset, comparing it with two state-of-the-art offline floorplan reconstruction methods: VecIM \cite{han2021vectorized} and RoomFormer \cite{yue2023connecting}. The former employs a gravity-aligned point cloud as input, while the latter uses a 2D density map obtained by projecting the point cloud along the gravity axis. In contrast, our method uses only stereo images as input. VecIM and RoomFormer both utilize a dense multi-view stereo (MVS) point cloud reconstructed by COLMAP \cite{schonberger2016structure}, which takes advantage of more comprehensive global optimization and additional multi-view information. This results in a more accurate point cloud than that generated by our SLAM system, ensuring fairness in the comparison. We uniformly sample 10K points from the reconstructed floorplan and compute the Hausdorff distance between these sampled points and the ground-truth point cloud to evaluate the reconstruction accuracy, using a dense, high-precision LiDAR point cloud as the ground truth.

\subsubsection{Evaluation on the VECtor Dataset}

Figure \ref{Vector_exp} shows the qualitative and quantitative results on the VECtor dataset, where our method achieves the lowest mean Hausdorff error and reconstructs a more regular floorplan closer to the real scene. Since VecIM directly applies efficient RANSAC to detect planes from the entire point cloud, it is highly susceptible to outliers and noise, which is an inevitable issue for MVS point clouds in weakly textured scenes. As a result, it extracts many incorrect planes, leading to numerous redundant layered structures. Moreover, VecIM enforces the floorplan to be closed, resulting in many erroneous closed regions. In contrast, although our method also faces challenges in weakly textured regions, it leverages plane points  from reliable plane landmarks that persist across multiple frames and undergo continuous optimization. While these points are sparser, they are robust matches carefully selected through strict consistency checks and outlier removal, resulting in significantly higher accuracy and structural regularity. Moreover, by incorporating trajectory information, our method correctly handles navigable areas and openings, avoiding erroneous closed regions.

\begin{figure*}[!t]
\centering
\includegraphics[width=0.75\textwidth]{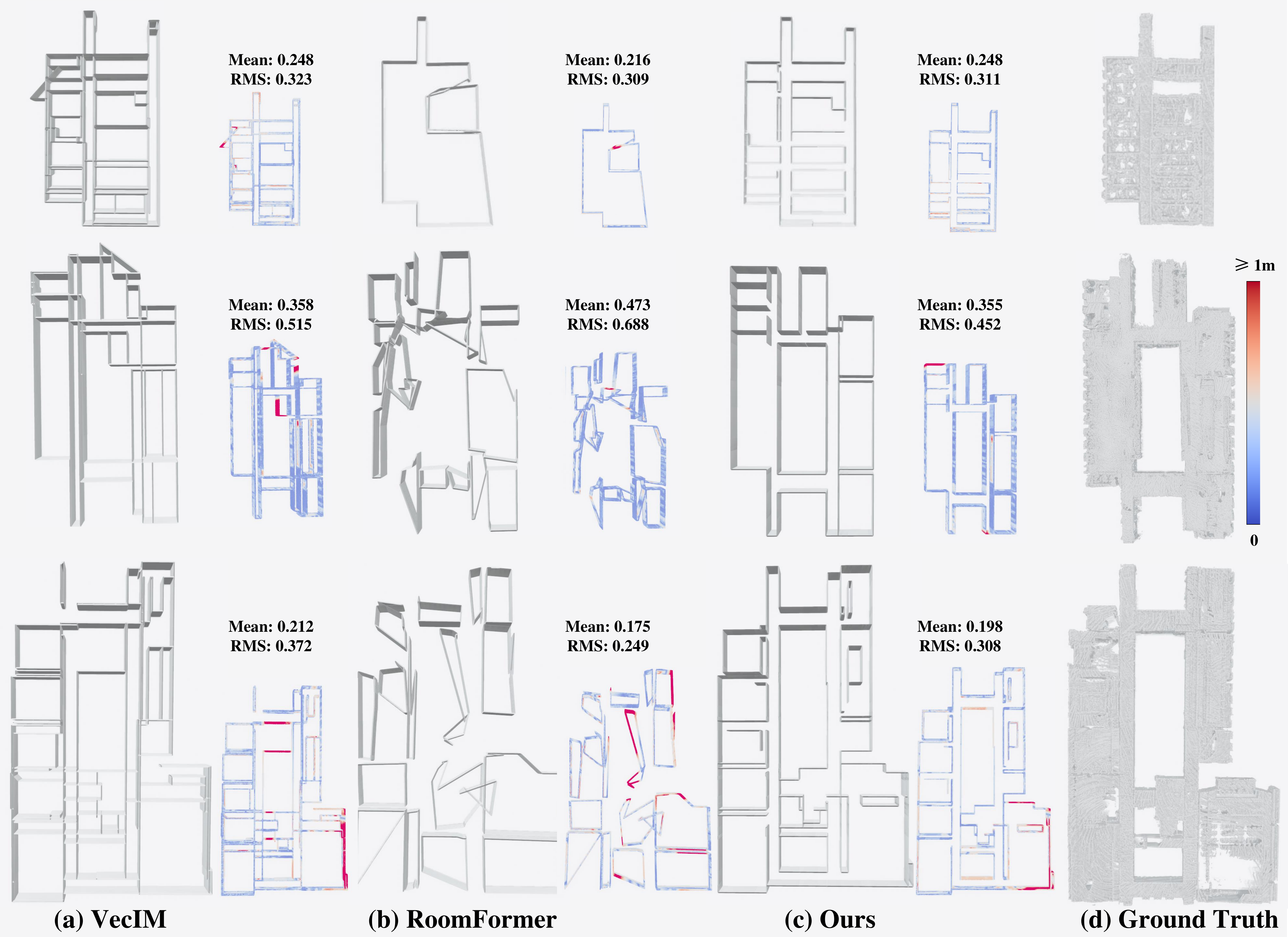}
\caption{Floorplan reconstruction results on the self-collected dataset of VecIM, RoomFormer and Ours.
For each reconstruction result, the mean Hausdorff error (m) and the error model are computed from the output to the ground truth. From top to bottom, we display the offices, café, and meeting rooms, respectively. Our method achieves the best performance in both reconstruction accuracy and regularity.}
\label{floorplan_reconstruction_performance_evaluation_self_collect}
\end{figure*}

\subsubsection{Evaluation on the Self-collected Dataset}

% The SelfCollect dataset comprises stereo images (640×320 resolution) captured using a Zed2 camera and LiDAR-captured point cloud data as ground truth. This dataset contains numerous non-lifelong structures, such as tables and chairs, which inherently complicate floorplan reconstruction. It includes three large-scale scenes (F3, F14, F17), each covering 1000 square meters. As demonstrated in Figure 2, we conducted comprehensive qualitative and quantitative comparisons of reconstruction results on this benchmark.
% Our method demonstrates superior reconstruction quality and achieves the highest metric accuracy. In contrast, RoomFormer, as a deep learning-based approach, exhibits limited generalization capability on the large-scale Self-collect dataset.
% VecIM is capable of reconstructing the main structure of the scene; however, due to the tendency of the stereo MVS pipeline to generate substantial noise at corridor corners, it struggles with noise in these areas and fails to properly handle openings at room entrances, especially in Floor3.
% In comparison, our proposed method effectively addresses these issues by incorporating the trajectory cross constraint.

We evaluate the floorplan reconstruction and long-term reconstruction performance of our method on the meeting rooms, offices, and café sequences from our self-collected dataset, all of which are complex indoor environments with numerous non-lifelong structures (e.g., tables, chairs), each spanning an area of approximately 1000 square meters. In the meeting rooms scenario, due to the inconsistent accessibility of different conference rooms, we conducted five separate data collections over two days, with each sub-session covering a different region of the scene, as shown in Fig. \ref{head_figure}. The qualitative and quantitative results of floorplan reconstruction on the self-collected dataset are presented in Fig. \ref{floorplan_reconstruction_performance_evaluation_self_collect}. Our method still reconstructs the most regular and faithful floorplan. Benefiting from rich textures, the MVS point cloud exhibits relatively low error, allowing VecIM to accurately reconstruct main structures. However, VecIM is unable to focus solely on extracting lifelong structures and detects numerous temporary structures such as table sides, adding redundant elements and failing to handle navigable areas effectively. In contrast, although our method may initially detect planes from non-lifelong objects, these planes typically do not persist across multiple frames and are not considered valid plane landmarks for the floorplan reconstruction module, enabling a focus on lifelong structures. Moreover, by fully leveraging trajectory information, we effectively handle openings in navigable areas, such as corridors and room entrances, ensuring that the reconstructed floorplan accurately represents passable regions. In addition, although our self-collected dataset contains room scenes, RoomFormer lacks generalizability for large-scale environments and thus can only generate a limited number of correctly identified rooms. Notably, by sequentially merging multiple subsequences, our method yields a complete large-scale floorplan that encompasses multiple meeting rooms, exhibiting remarkable accuracy and regularity, which strongly demonstrates the long-term globally consistent mapping capability of our approach. As a result, there is no need to acquire the complete scene data in a single pass, greatly reducing the complexity of data collection and reconstruction.

\subsection{Runtime Analysis}

\begin{table}[tp]
\centering
\begin{threeparttable}
\caption{Execution time comparison (in milliseconds).}
\label{threads_time_evaluation}
\begin{tabular}{|c|c|c|c|c|c|}
\hline
\multirow{2}{*}{Set.} & Sys. & \multicolumn{2}{c|}{ORB-SLAM3} & \multicolumn{2}{c|}{Ours} \\ \cline{2-6} 
 & Seq. & {\begin{tabular}[c]{@{}c@{}}school\\ 1224$\times$1024\end{tabular}} & {\begin{tabular}[c]{@{}c@{}}Meeting\\ 640$\times$360\end{tabular}} & {\begin{tabular}[c]{@{}c@{}}school\\ 1224$\times$1024\end{tabular}} & {\begin{tabular}[c]{@{}c@{}}Meeting\\ 640$\times$360\end{tabular}} \\ \hline \hline
\multirow{4}{*}{Tra.} & PE\textsuperscript{1} & 28.77 & 13.58 & 28.96 & 13.51 \\ \cline{2-6} 
 & PlE\textsuperscript{1} & - & - & 29.92 & 14.01 \\ \cline{2-6} 
 & PP\textsuperscript{1} & 8.44 & 5.54 & 11.2 & 8.32 \\ \cline{2-6}
 & Total & 37.36 & 19.28 & 41.38 & 22.61 \\ \hline \hline
 \multirow{3}{*}{Rec.} & SG\textsuperscript{1} & - & - & 42.30 & 37.37 \\ \cline{2-6} 
 & SS\textsuperscript{1} & - & - & 39.32 & 55.50 \\ \cline{2-6}
 & Total & - & - & 81.62 & 92.87 \\ \hline
\end{tabular}
\begin{tablenotes}
\footnotesize
\item[1] PE, PlE, PP, SG, and SS are the abbreviations for ORB Point Feature Extraction, Plane Feature Extraction, Pose Prediction, Segment Generation, and Segment Selection, respectively.
\end{tablenotes}
\end{threeparttable}
\end{table}

We evaluated our method on the school sequence from the VECtor dataset and on the meeting rooms sequence from our self-collected dataset, comparing the results with ORB-SLAM3. Table \ref{threads_time_evaluation} shows the runtimes of the two threads where our method and ORB-SLAM3 differ most; the other threads show comparable execution times. We extract Sobel and ORB features in parallel CPU threads, with the former being faster. Moreover, ORB extraction is the main time-consuming step compared to other plane extraction operations, such as mesh generation, clustering, and plane parameter estimation. As a result, our overall plane extraction time is close to that of ORB feature extraction. The floorplan reconstruction module comprises two primary components: segment generation and selection. We evaluate the runtime by measuring the duration of the final reconstruction in each sequence, which is the most time-consuming due to the largest number of segments. Notably, considering that valid plane landmarks are typically updated every several frames, we perform floorplan reconstruction every five frames. This frequency is sufficient to provide timely updates, thereby enabling real-time floorplan reconstruction.
We also recorded the total reconstruction time consumed by VecIM, including the MVS dense point cloud reconstruction and the floorplan reconstruction stages. The first stage took 16 hours and 38.9 minutes, and the second stage took 5.1 minutes, resulting in a total of 16 hours and 44 minutes. In contrast, we processed only stereo images at 45 FPS, and completed the entire floorplan reconstruction in 9.4 minutes, which is two orders of magnitude faster than VecIM. This result strongly demonstrates the efficiency and practicality of our method.

\section{Conclusion}
% In this paper, we propose Floorplan-SLAM, a real-time floorplan reconstruction system that only uses a stereo camera. Unlike existing methods that depend on pre-acquired dense point clouds or computationally expensive learning-based methods, Floorplan-SLAM achieves incremental floorplan reconstruction through a tightly coupled point-plane-based stereo SLAM framework. By integrating trajectory constraints and plane regularization, our approach ensures high accuracy, robustness, and adaptability in complex indoor environments. Experimental results on both the VECtor and Self-Collect datasets demonstrate that Floorplan-SLAM outperforms state-of-the-art methods, achieving real-time, high-fidelity floorplan reconstruction while surpassing existing stereo SLAM approaches in accuracy and completeness.

% Our approach addresses the critical challenges of existing methods, such as the reliance on expensive sensors, high computational costs, and time-consuming offline processing. By leveraging an innovative plane extraction algorithm and a tightly coupled floorplan reconstruction module within the multi-session SLAM framework, we

This paper presents Floorplan-SLAM, a novel approach for real-time, high-accuracy, and long-term floorplan reconstruction, significantly reducing reconstruction time while maintaining superior accuracy and robustness. The proposed method consistently outperformed state-of-the-art approaches in three key aspects: plane extraction robustness, pose estimation accuracy, and floorplan reconstruction fidelity and speed. Experimental results on the VECtor dataset and our self-collected dataset demonstrate the effectiveness of Floorplan-SLAM. Specifically, our algorithm is able to reconstruct a 1000~$\mathrm{m}^{2}$ floorplan in just 9.4 minutes, compared with the over 16 hours and 44 minutes required by the existing offline method \cite{han2021vectorized}, achieving a speedup of two orders of magnitude, while yielding more accurate and structured reconstruction results. The real-time performance, achieving 25--45 FPS without GPU acceleration, further underscores the efficiency and practicality of our approach for large-scale indoor floorplan reconstruction. Future work will focus on deeply integrating the floorplan with the SLAM system to fully leverage the high-level scene understanding it provides, thereby enabling hierarchical localization and map merging in challenging scenarios.

% Additionally, Floorplan-SLAM demonstrated remarkable robustness in weakly textured scenes, thanks to the effective integration of spatially complementary features in the plane extraction process. The map merging capability of multi-session SLAM further enabled long-term floorplan reconstruction without redundant data collection, providing a solid foundation for continuous indoor navigation and scene understanding.

% In summary, Floorplan-SLAM offers a significant advancement in the field of real-time indoor floorplan reconstruction, providing both practical and scalable solutions for robotic navigation in real-world environments. Future work will focus on further improving the system’s scalability and robustness in increasingly complex environments.

% Future work will focus on further integrating the floorplan with the SLAM system to achieve applications such as hierarchical localization and map merging.

% \addtolength{\textheight}{-12cm}

\bibliographystyle{IEEEtran}
\footnotesize
\bibliography{IEEEabrv,reference}

\end{document}